\documentclass[a4paper,twoside]{article}

\usepackage{apalike}
\let\bibhang\relax
\usepackage{natbib}
\usepackage{epsfig}
\usepackage{subcaption}
\usepackage{calc}
\usepackage{amssymb}
\usepackage{amstext}
\usepackage{amsmath}
\usepackage{amsthm}
\usepackage{multicol}
\usepackage{pslatex}
\usepackage{algorithm2e}
\usepackage[bottom]{footmisc}
\usepackage{url}
\usepackage{SCITEPRESS}

\begin{document}

\title{Secure Visual Data Processing via Federated Learning}

\author{\authorname{Pedro Santos\sup{1}\orcidAuthor{0009-0002-8601-7905}, Tânia Carvalho\sup{1}\orcidAuthor{0000-0002-7700-1955}, Filipe Magalhães\sup{2} and Luís Antunes\sup{1,2}\orcidAuthor{0000-0002-9988-594X}}
\affiliation{\sup{1}Faculty of Sciences, University of Porto}
\affiliation{\sup{2}TekPrivacy, Porto}\email{up201907254@fc.up.pt}}


\abstract{As the demand for privacy in visual data management grows, safeguarding sensitive information has become a critical challenge. 
This paper addresses the need for privacy-preserving solutions in large-scale visual data processing by leveraging federated learning. Although there have been developments in this field, previous research has mainly focused 
on integrating object detection with either anonymization or federated learning. However, these pairs often fail to address complex privacy concerns. On the one hand, object detection with anonymization alone can be vulnerable to reverse techniques.
On the other hand, federated learning may not provide sufficient privacy guarantees.
Therefore, we propose a new approach that combines object detection, federated learning and anonymization. 
Combining these three components aims to offer a robust privacy protection strategy by addressing different vulnerabilities in visual data.
Our solution is evaluated against traditional centralized models, showing that while there is a slight trade-off in accuracy, the privacy benefits are substantial, making it well-suited for privacy sensitive applications.
}

\onecolumn \maketitle \normalsize \setcounter{footnote}{0} \vfill

\section{\uppercase{Introduction}}
\label{sec:introduction}

The exponential growth of visual data has raised numerous privacy concerns related to sensitive information embedded within the content. These concerns are particularly evident in healthcare, surveillance, social media and autonomous vehicles, where large amounts of personal and identifiable data are generated and processed. Many regulatory frameworks have been created to protect sensitive information, namely, the General Data Protection Regulation (GDPR) is the standard privacy regulation that has influenced data protection laws worldwide. This regulation requires that personal data should be processed in a way that ensures its privacy, particularly when handling data that may contain identifiable information.

However, conventional solutions for managing visual information focus on applying machine learning approaches in centralized processing. This centralization involves sending and storing the data in a single location for further analysis, which often exposes the sensitive information in these systems to unauthorized access~\citep{10.1145/2810103.2813687}. Additionally, even considering anonymization techniques, these may be insufficient, particularly in terms of their inability to prevent sophisticated attacks such as model inversion and adversarial attacks, which can expose sensitive information from images and videos despite anonymization or encryption efforts~\citep{fredrikson2015model,goodfellow2014explaining}.
These data breaches may cause severe consequences for individuals and financial losses for organizations. 
Besides the privacy threats in data centralization, the vast quantity of visual data renders manual processing impractical, creating scalability challenges in real-time applications, which traditional approaches cannot adequately address~\citep{bharati2022federated}.

Federated Learning (FL)~\citep{mcmahan2023communicationefficientlearningdeepnetworks} emerges as a compelling approach to tackling these challenges. By enabling the training of Artificial Intelligence (AI) models across decentralized data sources without the transfer of the actual data, FL inherently enhances privacy. Thus, we aim to leverage AI for visual data labelling within a FL framework.

Despite the recent advances in FL, the integration of FL with AI-driven visual data management tools, particularly those focused on labelling and anonymization, remains under-explored. 
Namely, existing research tends to focus on either anonymization techniques~\citep{andrade2024privacy} or federated learning for privacy preservation~\citep{yu2019federated}, but rarely addresses the integration of both, let alone the inclusion of visual data labelling as a key component. Most studies emphasize anonymization to protect identity in images or federated learning to decentralize data, but fail to consider a unified approach that tackles all three aspects. This gap leaves room for privacy vulnerabilities, especially in scenarios where sensitive visual data must be shared or processed across multiple devices or organizations.

To our knowledge, this paper presents the first approach combining these three components: visual data labelling, federated learning and anonymization. Object detection identifies sensitive visual regions such as faces and license plates while FL avoids the need to share raw data, and then, the detected critical regions are effectively masked through anonymization techniques, resulting in a multi-layered defence against privacy breaches. 
Our key contributions are highlighted as follows.
\begin{itemize}
    \item \textbf{Visual data labelling in a cutting edge FL framework:} we use a recent and straightforward FL technology with a well-known object detection algorithm for an efficient and secure visual data labelling system.
    \item \textbf{Anonymization layer:} we apply obfuscation techniques to anonymize sensitive visual data in the federated environment.
    \item \textbf{Performance evaluation:} we conduct comprehensive experiments to evaluate the performance and scalability of the proposed solution. 
\end{itemize}
This improved approach promises to balance the need for visual data utility with elevated privacy requirements, paving the way for advanced and secure visual information management.

The remainder of the paper is organised as follows: Section~\ref{sec:literaturereview} provides a literature review and discussion of related work concerning AI-driven visual data management, image anonymization techniques, and FL. Section~\ref{sec:methodology} outlines the research design, including the selection of datasets, the implementation of AI algorithms, and the FL framework.
Section~\ref{sec:results} presents the results followed by a thorough discussion on the implications of our findings for visual data management. Section~\ref{sec:conclusion} provides the conclusions and highlights potential avenues for future research.

\section{\uppercase{Literature Review}}
\label{sec:literaturereview}
This paper explores three main domains: object detection, anonymization techniques and federated learning (FL). In this section, we present a comprehensive review of each area, alongside our contributions
to the current state-of-the-art.

\subsection{Object Detection}
Object detection remains a key focus of research, with many algorithms tested across popular datasets.

Many surveys stress the importance of features such as color, texture, shape and spatial relationships for a more effective visual content labelling process in tasks like image retrieval, annotation, and object detection~\citep{Veltkamp2000,9770283,Bouchakwa2020}.
There is a trend in self-supervised learning methods since they can be used on unlabeled data, which is advantageous for scaling up computer vision applications~\citep{9770283}. Despite all the advances in object detection, researchers still point to ongoing challenges in achieving high labelling accuracy and efficiency~\citep{zou2023object}. Automated methods are improving, but manual and semi-automated labelling still plays an important role, especially in ensuring quality in critical applications~\citep{girshick2014rich}.

More sophisticated approaches include deep learning methods that leverage neural networks for automatic feature extraction and object recognition.~\citet{zou2023object} classify deep learning-based object detection algorithms into two main categories: one-stage and two-stage algorithms.

\paragraph{One-Stage.}
These type of algorithms are widely used in object detection because they offer faster detection speeds, making them suitable for real-time tasks.
These detectors aim to directly predict object bounding boxes and class labels in a single forward pass, without the need for a two-stage process involving region proposals, as seen in two-stage detectors.

Multiple studies have been conducted testing one-stage algorithms in real-world applications. For example,~\citet{zhu2022} explore the importance of accurate and efficient traffic sign detection and recognition in Intelligent Transportation Systems. The authors also show that YOLOv5~\citep{jocher2020ultralytics} is more accurate and faster than SSD~\citep{liu2016ssd} algorithm for traffic sign recognition. 
Additionally, many other one-stage algorithms, such as RetinaNet~\citep{lin2017focal}, DETR~\citep{carion2020end}, CornerNet~\citep{law2018cornernet}, CenterNet ~\citep{duan2019centernet} offer unique advanced techniques for object detection in real-world scenarios. These models employ innovations like focal loss, transformers, and keypoint detection. 

\paragraph{Two-Stage.}
As the name suggests, these type of algorithms works by splitting the process into two different stages, one is used to determine interest regions where objects are located, and the other to classify and further refine the localization of the object.

R-CNN~\citep{girshick2014rich} was one of the first approaches to deep-learning-based object detection algorithms of this category. Its successors, Fast RCNN~\citep{girshick2015fast} and Faster RCNN~\citep{ren2015faster}, brought significant speed and accuracy improvements. In particular, Faster RCNN remains highly competitive due to its innovative Region Proposal Network, which streamlines the object proposal process. Later advancements like SPPnet~\citep{he2015spatial} introduced spatial pyramid pooling for more efficient multi-scale processing, while FPN~\citep{lin2017feature} and Mask RCNN~\citep{he2017mask} refined detection capabilities further. HTC (Hybrid Task Cascade)~\citep{chen2019hybrid} built on these innovations, offering improvements in both object detection and instance segmentation.

Table~\ref{tab:stages} presents a summary of one and two-stage algorithms, highlighting their distinct advantages and common applications. 

\begin{table*}[t]
\caption{Comparison of one-stage and two-stage object detection algorithms.}
\centering
\scriptsize
\begin{tabular}{|c|c|c|c|}
\hline
\textbf{Types} & \textbf{Algorithms} & \textbf{General Advantages} & \textbf{Common Use Cases} \\ 
\hline
One-Stage & 
\begin{tabular}[c]{@{}l@{}} 
- YOLO ~\citep{redmon2016you}\\ 
- SSD ~\citep{liu2016ssd}\\ 
- RetinaNet ~\citep{lin2017focal}\\ 
- DETR ~\citep{carion2020end}\\ 
- CornerNet ~\citep{law2018cornernet}\\ 
- CenterNet ~\citep{duan2019centernet}
\end{tabular} & 
\begin{tabular}[c]{@{}l@{}} 
- Faster Object Recognition\\ 
- Memory Efficiency during Inference\\ 
- Suitable for Real-Time Applications\\ 
- Require fewer Computation resources
\end{tabular} & 
\begin{tabular}[c]{@{}l@{}} 
- Autonomous Vehicles\\ 
- Real-Time Surveillance Systems\\ 
- Augmented Reality
\end{tabular} \\ 
\hline
Two-Stage & 
\begin{tabular}[c]{@{}l@{}} 
- R-CNN ~\citep{girshick2014rich}\\ 
- Fast R-CNN ~\citep{girshick2015fast}\\ 
- Faster R-CNN ~\citep{ren2015faster}\\ 
- FPN ~\citep{lin2017feature}\\ 
- SPPNet ~\citep{he2015spatial}\\ 
- HTC ~\citep{chen2019hybrid}
\end{tabular} & 
\begin{tabular}[c]{@{}l@{}} 
- Better Accuracy\\ 
- Adaptability to Different Tasks\\ 
- Reduced False Positives\\ 
- Work better in complex/challenging scenes
\end{tabular} & 
\begin{tabular}[c]{@{}l@{}} 
- Medical Imaging Applications\\ 
- Object Detection in Satellite Imagery\\ 
- Industrial Quality Control
\end{tabular} \\ 
\hline
\end{tabular}
\label{tab:stages}
\end{table*}

\subsection{Anonymization Techniques}

Anonymization has gained significant attention in recent years, driven by the increase in the number of privacy regulations and the rapid growth in the capabilities of AI-based applications. 
This topic has been extensively surveyed by many researchers~\cite{senior2009protecting,amsterdam_intelligence,ren2018learning}. Key findings indicate that most existent anonymization techniques inherently create a trade-off between privacy and the utility of the final image.  
While recent efforts, such as those discussed by~\citet{amsterdam_intelligence}, have been made to create a solution that would achieve a better compromise in terms of image usability, it is still unclear whether or not an attacker would be able to successfully revert these operations with sufficient resources and expertise. Also, the choice of the right technique to be used is highly dependent on the specific application, data type, and desired level of privacy. The surveys emphasize the need for careful consideration of these factors to choose the most appropriate method.

To address these limitations, more advanced techniques have been developed and adopted. For example, Generative Adversarial Networks (GANs) can modify sensitive features by replacing them with realistic generated alternatives, ensuring a higher level of privacy while preserving the overall context of the visual scene.~\citet{todt2022fant} explore various face anonymization techniques, including GANs, blurring, and pixelation, evaluating their effectiveness in preventing de-anonymization or identity reconstruction.

Recently,~\citet{hukkelaas2023does} compare different anonymization techniques, including traditional methods (blurring and masking) and realistic methods (generating synthetic faces and bodies).
The authors reveal that realistic anonymization can significantly reduce performance degradation caused by traditional methods, especially for face anonymization. Despite these advancements, a significant challenge lies in finding the right balance between anonymization and data utility. Also, more advanced techniques tend to be computationally expensive, making it essential to take into account when choosing the best approach for the targeted application. 

\subsection{Federated Learning}

FL has emerged as a promising approach to address the challenges of privacy and data ownership that have surfaced with the growing collection of data. This approach enables participating entities to train a shared model collaboratively without sharing the actual data. Instead, participants only share model updates to a central server that aggregates all the received information. Thereby, FL aims to mitigate the risks of data breaches and preserves the privacy of individual users~\citep{guan2024federatedlearningmedicalimage}. This approach has been employed in domains such as medical imaging~\citep{kaissis2020secure}, finance~\citep{long2021federatedlearningopenbanking}, and autonomous vehicles~\citep{pokhrel2020federated}.
The two most common FL settings include Horizontal (HFL) and Vertical Federated Learning (VFL).

\paragraph{Horizontal Federated Learning.}

HFL is used when multiple participants have datasets with the same feature space but different sets of individuals. For example, several hospitals, each with their patient records with similar attributes, such as blood test results or medical images, can use HFL to collaboratively train a shared model. Each hospital trains a local model on its dataset and only transmits the learned model parameters to a central server, ensuring that sensitive data is not shared.

~\citet{9874186} explore various aggregation methods, privacy techniques, and system architectures, providing a comprehensive overview of HFL methodologies. The authors present addressing measures for data heterogeneity and fairness among participants.
HFL has also been applied to recommendation systems~\citep{wang2024horizontal}. The authors explore the use of matrix factorization and federated collaborative filtering, aiming to enhance recommendation quality while preserving user privacy. Similarly, in the domain of computer vision, HFL has been employed for tasks like object detection and image segmentation~\citep{Mandal_2024}. This approach leverages federated models such as Faster R-CNN and Fully Convolutional Networks to handle the growing volume of visual data.

\paragraph{Vertical Federated Learning.}
The application of VFL involves participants with datasets that share a common set of individuals but contain different features for these individuals. For instance, different hospitals might have overlapping patients, but while one might have access to demographic information, the other might have genetic or medical imaging data. 

Finding the common individuals that are present in each organization's dataset requires the application of Private Set Intersection (PSI)~\citep{angelou2020asymmetricprivatesetintersection},  a crucial component of any VFL system. PSI is a cryptographic technique designed to detect common data across two or more participants without disclosing any other information beyond the shared records.

Despite being a relatively new approach in federated learning, VFL is gaining much attention in the relevant literature. In particular, new surveys have been published~\citep{yang2023survey,liu2024vertical,yu2024survey}, pointing out the possibilities within the field when combining specialized hardware with advanced privacy-preserving techniques, and optimized training protocols. These studies present techniques like homomorphic encryption, secure multi-party computation, and differential privacy for maintaining the confidentiality of sensitive information. Moreover, the surveys discuss the use of training protocols specifically designed to address challenges such as communication overhead, data heterogeneity, and model convergence, making VFL a more viable and robust solution for real-world applications.

\subsection{Current Research Directions}

In the relevant literature, various efforts have been made to integrate object detection with either anonymization or FL.
On the one hand,~\citet{andrade2024privacy} analyzes face detection using anonymization techniques such as blurring and pixelation, highlighting the challenge of balancing privacy and utility. On the other hand,~\citet{yu2019federated} proposes using FL for object detection but does not address anonymization, leaving potential privacy risks unresolved.
Similarly,~\cite{memia2023federated} apply FL to real-time object detection using YOLOv8~\citep{Jocher_Ultralytics_YOLO_2023}, focusing on training models across edge devices without sharing raw data. This work effectively demonstrates the use of FL for decentralized object detection. However, while it successfully mitigates data-sharing risks, it does not integrate anonymization techniques, leaving sensitive visual data potentially exposed to privacy threats. Our approach builds on this work by adding an anonymization layer to further protect sensitive information, offering a more comprehensive privacy-preserving solution.

Therefore, we aim to employ an approach that boosts privacy by combining object detection, federated learning and anonymization. To the best of our knowledge, we are the first to combine these three components in which besides leveraging an AI model through FL, we apply an additional anonymization layer on the detected sensitive visual data.

\section{\uppercase{Visual Data Protection and Labelling}}
\label{sec:methodology}

This section outlines the experimental evaluation procedure. We detail the methodology, providing an overview of the high-level architecture of the proposed system, including the algorithm used to train the model and a description of the dataset.

\subsection{Methodology} 

\begin{figure}[!htb]
\centering
    \includegraphics[width=\linewidth]{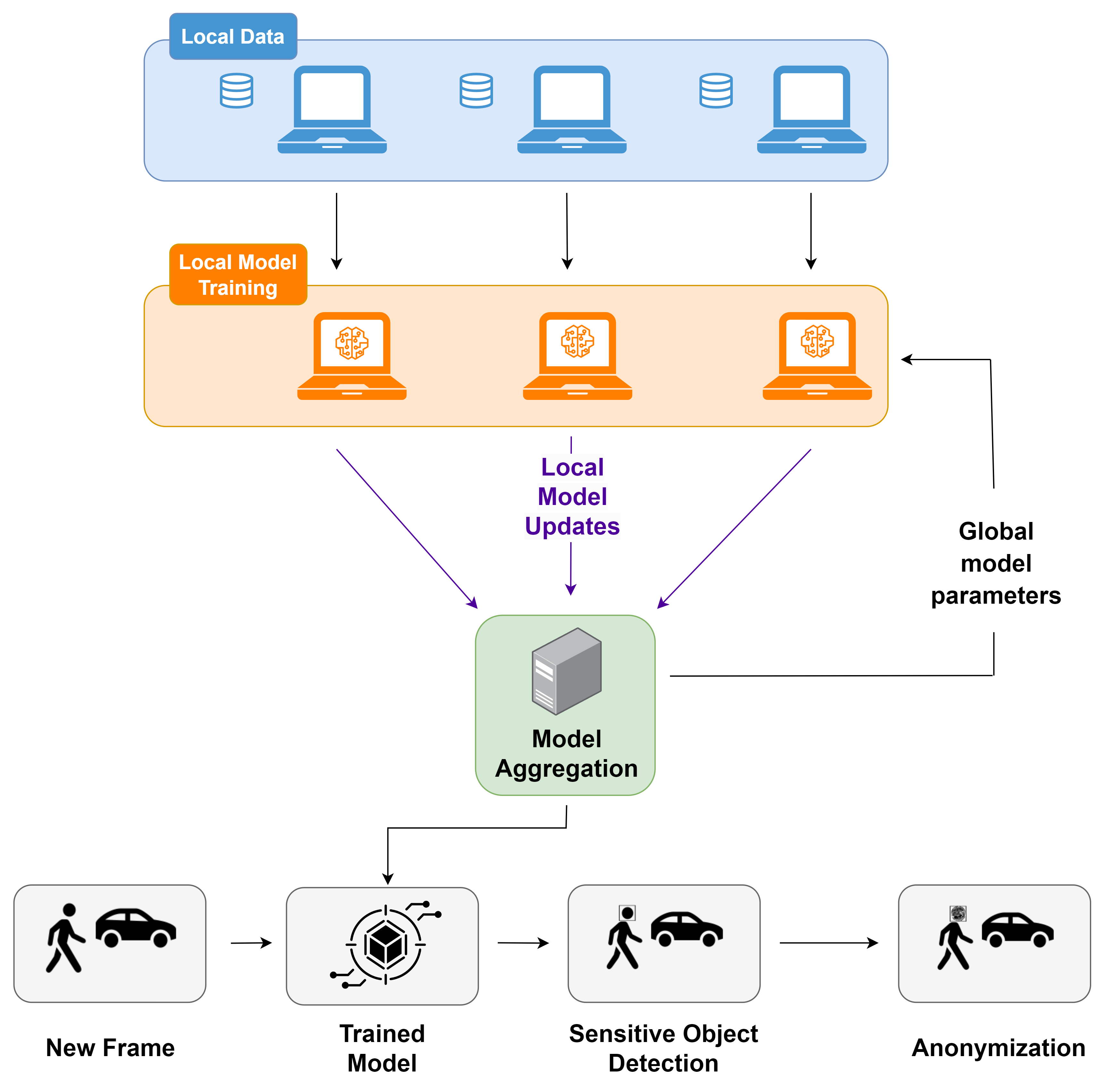}
    \caption{Methodology for sensitive data detection and anonymization using federated learning.}
    \label{fig:methodology}
\end{figure}

We aim to develop a robust framework that ensures data privacy, facilitates accurate object detection, and maintains the integrity of the anonymized data for analysis. As such, we present our proposed methodology in Figure~\ref{fig:methodology}~\footnote{Icons: “Object Detection” by Edward Boatman \url{https://thenounproject.com/icon/object-detection-6109597/} and “Object Detection” by Nanang A Pratama \url{https://thenounproject.com/icon/object-detection-6943616/} from Noun Project.}.

The overall process is divided into several steps:

\begin{itemize}
    \item \textbf{Local model training}: in the first step, each participant trains a local object detection model on their private visual data.
    \item \textbf{Local model updates transmission}: participants send only the model updates (e.g. learned weights or gradient) to a central server after training.
    \item \textbf{Model aggregation}: the central server aggregates the updates from all participants to refine a global model. Federated averaging is often used to combine local models iteratively, improving the global model's performance at each training round.
    \item \textbf{Deployment of the trained model}: once the global model reaches satisfactory performance, it can be deployed for tasks like processing new images or video frames in real time, benefiting from the collective knowledge gathered during training.
    \item \textbf{Sensitive data detection}: the deployed model uses object detection algorithms to identify and label targeted classes in visual data.
    \item \textbf{Anonymization}:
    finally, after detection, the sensitive data is anonymized.
\end{itemize}

In summary, this system utilizes FL to train the model across multiple devices without centralizing the data. The object detection component identifies and locates sensitive areas within the visual content, such as faces or license plates. To ensure that privacy is maintained, an anonymization process will then be applied to these sensitive areas, masking or altering them to prevent identification. We aim to create a scalable and secure solution that can be used in various applications, from surveillance and healthcare to social media and content-sharing platforms, where privacy and data protection are fundamental. 

\subsection{Methods}

Concerning each component of our methodology, we employ FL with Flower~\citep{beutel2020flower}, object detection using YOLOv8~\citep{Jocher_Ultralytics_YOLO_2023}, and Gaussian blur for anonymization.

Among the several object detection methods, YOLO is particularly relevant due to its superior detection speed when compared to other popular algorithms making it better suited for real-time applications. Also, the algorithm has undergone continuous improvements, enhancing both its efficiency and accuracy over successive versions, further solidifying its importance for tasks that require quick and accurate object detection. Thus, we use YOLOv8~\citep{Jocher_Ultralytics_YOLO_2023}, which demonstrates strong performance across a wide range of datasets and scenarios, improving its generalization capabilities.
This adaptability is particularly relevant as our system is designed to process diverse visual content. Furthermore, the efficiency offered by YOLOv8 minimizes computational demands on edge devices, making the system more accessible in distributed environments that do not require high-end hardware.

However, integrating YOLOv8 in an FL system may pose technical challenges given the under-exploration of their combination. In contrast to~\citet{memia2023federated} that uses the FEDn framework, we use the Flower~\citep{beutel2020flower} with YOLOv8 to train federated models. Flower offers greater flexibility and compatibility with various machine learning libraries, including PyTorch and TensorFlow.
Its efficient communication protocols reduce data transfer needs, while customizable participant selection optimizes the training process. The tool’s modular architecture also simplifies integration into our system, making Flower an ideal choice for our FL setup. 
This compatibility with YOLOv8 allows efficient training across distributed devices. Finally, the extensive documentation, active community support, and continuous updates offered by Flower ensure that any challenges encountered during the implementation of our system were effectively addressed.

\subsection{Data}

We collected images from the Open Images Dataset V6~\cite{OpenImages2}~\footnote{Publicly accessible at \url{https://storage.googleapis.com/openimages/web/visualizer/index.html}}, focusing on images annotated with "Vehicle registration plate" and "Human face" corresponding to the targeted classes for our study  We made sure to convert the annotations into the appropriate format, normalizing bounding box coordinates based on image dimensions, thereby ensuring compatibility with YOLOv8.

Our final dataset comprises 29.690 images, divided into training, validation, and testing sets:
 \begin{itemize}
     \item \textbf{Training set}: 14.485 images (48.8\%)
     \item \textbf{Validation set}: 3.848 images (13.0\%)
     \item \textbf{Testing set}: 11.357 images  (38.2\%)
 \end{itemize}

This distribution ensures a balanced evaluation of the model's performance for our use case.

The FL setup involves a distributed learning process across multiple participants, each having access to a portion of the overall dataset. For this experiment, we limit the number of participants to three. This decision was made to control the complexity of the experiment and to focus on understanding the impact of data distribution among a small number of participants.
Also, each participant has a distinct partition of the dataset, ensuring no single participant has access to the entire dataset. This setup allows us to simulate a realistic FL scenario where data is distributed across different nodes but with the added constraint of limited data availability per participant.

\section{\uppercase{Results}}
\label{sec:results}

This section presents the experimental results of our proposed methodology. First, we establish a baseline using a centralized model, i.e. without Federated Learning (FL), for comparison purposes. Then, we evaluate the effects of increasing the number of training rounds on the federated model's performance. We also compare different aggregation methods in FL, analyzing their impact on runtime and model performance. Finally, we examine the effectiveness of the anonymization layer in preserving privacy while maintaining the integrity of the detected objects. This comprehensive evaluation aims to provide insights into the efficiency and practicality of deploying FL for sensitive data detection and anonymization.

The results were obtained using a high-performance virtual machine equipped with a 16-core CPU, 128 GB of RAM, and an NVIDIA GeForce RTX 3090 GPU (24 GB VRAM) to train and validate the YOLOv8 model. This setup ensured sufficient computational power to train the model with a large-scale dataset.

\subsection{Baseline Model}

To establish a baseline, we train the YOLOv8 model on the entire dataset using a centralized approach, where all data is available on a single system. Standard hyperparameters are used to minimize the loss functions related to bounding box regression, class prediction, and distribution focal loss.
To evaluate the performance of the trained model, we use widely recognized metrics such as precision, recall, and mean average precision (mAP). Precision measures the ratio of true positive detections to the total number of positive detections, indicating the model's accuracy in identifying relevant objects. Recall measures the ratio of true positive detections to the total number of actual positives, reflecting the model's ability to detect all relevant objects. The mAP combines precision and recall, evaluated at different intersections over union thresholds such as mAP50 and mAP50-95. 

\begin{figure}[!htb]
\centering
    \includegraphics[width=1\linewidth]{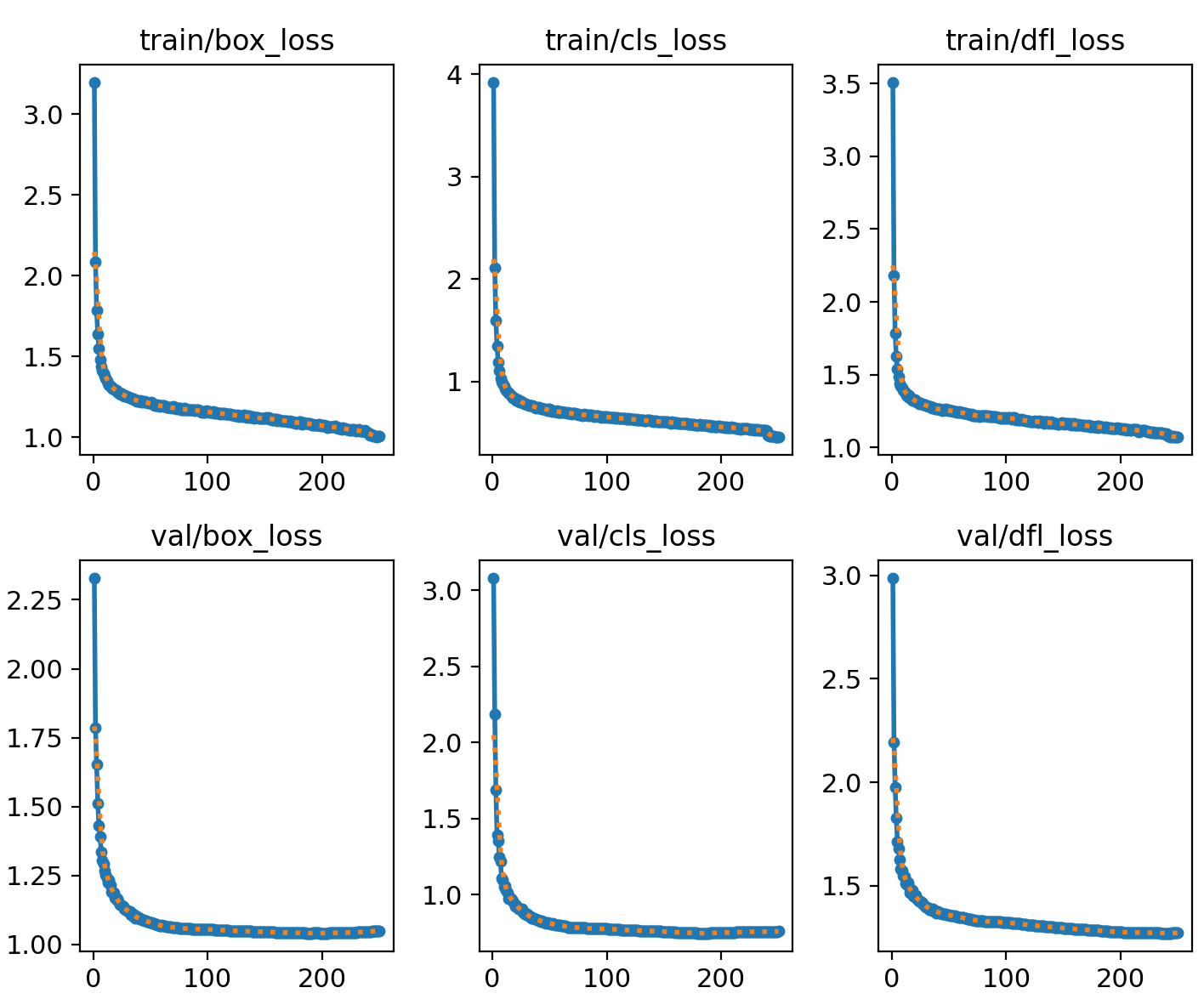}
    \caption{Train and validation loss from baseline YOLOv8.}
    \label{images:baseline}
\end{figure}

Figure~\ref{images:baseline} shows the training and validation loss curves of the baseline model for demonstrative purposes -- we aim to visualize the model's behavior.
We observe a consistent decrease in both training and validation losses over the epochs, indicating effective learning and model optimization.

Table~\ref{tab:epoch_variation_baseline} presents the results regarding the test set. 
The model's performance improves as the number of epochs increases, reaching its maximum at 200 epochs. 
This trend indicates that the model was successfully learning and optimizing over time. However, we observe that using 100 epochs, the centralized model provides lower loss, but the remaining metrics are generally slightly worse than training the model with 150 epochs.

\begin{table*}[htb]
\caption{Performance metrics for the baseline model (centralized) across varying epochs.}\label{tab:epoch_variation_baseline} 
\centering
\scriptsize
\begin{tabular}{|c|c|c|c|c|c|c|}
\hline
Epochs & mAP50 & mAP50-95 & Recall & Precision & Loss & Training Time (sec)\\
\hline
25     & 76.07\%  & 51.21\%  & 72.21\% & 81.00\%  & 0.00039 & 1,740\\
\hline
50     & 78.52\%  & 54.01\%  & 75.46\% & 83.20\% & 0.00039 & 3,357 \\
\hline
100    & 79.41\%  & 55.25\%  & 76.44\% & 82.01\% & 0.00036 & 6,421\\
\hline
150    & 79.69\%  & 55.36\%  & 76.43\% & 84.21\% & 0.00038 & 9,910\\
\hline
\underline{200}    & \underline{80.05\%}  & \underline{56.11\%}  & \underline{77.34\%} & \underline{84.32\%} & \underline{0.00037} & \underline{12,760}\\
\hline
\end{tabular}
\end{table*}

\subsection{Federated Learning Setting}
We now evaluate the performance of the proposed FL framework by analyzing key factors such as training epochs, aggregation methods, and communication costs in comparison with the baseline. 
\subsubsection{Epoch variation on model performance}

The number of epochs represents a critical hyperparameter in machine learning, as it determines the number of times the learning algorithm will work through the entire training dataset. Typically, increasing the number of epochs can lead to enhanced model performance, as it allows the model more opportunities to adjust its parameters and reduce errors. The baseline results confirm this. However, in FL, increasing the number of epochs may result in minimal performance gains while considerably extending training time due to communications in this procedure. 
Table~\ref{tab:epoch_variation} shows the results using the same epoch variation as previously and the aggregation method FedAvg~\citep{mcmahan2017communication}. 

\begin{table*}[ht]
\caption{Performance metrics results for the federated model with 5 rounds across varying epochs.}
\label{tab:epoch_variation}
\centering
\scriptsize
\begin{tabular}{|c|c|c|c|c|c|c|}
\hline
Epochs & mAP50 & mAP50-95 & Recall & Precision & Loss & Training Time (sec) \\
\hline
25     & 62.55\%  & 43.79\%  & 48.71\% & 75.03\%   & 0.00040 & 8,047 \\
\hline
50     & 56.23\%  & 38.69\%  & 51.09\% & 67.53\%   & 0.00042 & 15,606 \\
\hline
100    & 63.77\%  & 44.85\%  & 50.84\% & 78.85\%   & 0.00040 & 30,359 \\
\hline
150    & 64.70\%  & 45.01\%  & 58.88\% & 72.12\%   & 0.00040 & 38,183 \\
\hline
\underline{200}    & \underline{74.67\%}  & \underline{52.32\%}  & \underline{68.52\%} & \underline{77.81\%}   & \underline{0.00038} & \underline{49,633} \\
\hline
\end{tabular}
\end{table*}

In general, we verify the same trend as before: the higher the number of epochs, the better the performance of the federated model. This is particularly evident in mAP50, mAP50-95 and recall, which compared to the baseline, these metrics present higher differences between the 25 and 200 epochs. Nevertheless, we observe some fluctuations, especially in precision with an increase of less than 3\% between 25 epochs and 200 epochs. Also, the metric loss presents higher fluctuations. These outcomes may be attributed to the communication and aggregation model in FL. 
Despite the reached peak at 200 epochs, the federated model generally underperforms the baseline. This may be attributed to the distributed nature of data in FL, which may hinder its ability to generalize as effectively as the baseline model. 
Consequently, the run time for 200 epochs is four times longer than the baseline.
Therefore, it is essential to assess the trade-off between the utility gains over higher epochs and the computational resources required to achieve them, making it crucial to tailor the choice of epochs to the specific needs and constraints of the targeted application.

\subsubsection{After-effect of adding more rounds}
To further analyze the impact of increasing the number of communication rounds, we conduct experiments with the number of rounds set to 3, 4, 5, 6, 7, and 8. We set the number of training epochs at 200 for this scenario given the good results achieved previously. This allows us to observe how the model's performance evolves with successive rounds of communication between the edge devices and the central server. Table~\ref{tab:evolution_rounds} summarizes the results.%
\begin{table*}[ht]
\caption{Performance metrics results for the federated model with 200 epochs over different rounds.}
\label{tab:evolution_rounds}
\centering
\scriptsize
\begin{tabular}{|c|c|c|c|c|c|c|}
  \hline
  Round & mAP50 & mAP50-95 & Recall & Precision & Loss & Training Time (sec) \\
  \hline
  3 & 28.84\% & 19.21\% & 24.30\% & 36.17\%& 0.00040 & 32,834 \\
  \hline
  4 & 56.41\% & 40.95\% & 50.73\% & 76.77\% & 0.00039 & 43,386 \\
  \hline
  5 & 74.68\% & 52.32\% & 68.52\% & 77.82\% & 0.00042 & 49,633 \\
  \hline
  6 & 74.42\% & 52.29\% & 68.09\% & 78.76\% & 0.00040 & 60,545 \\
  \hline
  7 & 76.01\% & 53.44\% & 71.09\% & 77.75\% & 0.00039 & 61,601 \\
  \hline
  \underline{8} & \underline{76.17\%} & \underline{53.58\%} & \underline{71.27\%} & \underline{78.19\%} & \underline{0.00037} & \underline{67,797} \\
  \hline
\end{tabular}
\end{table*}

Increasing the number of rounds generally leads to improved performance results.
We note that a lower number of rounds provides a reduced ability of the model to detect relevant features. Namely, 3 rounds show considerably lower results than 5 rounds -- the majority of the metrics show an increase greater than 40\% between 3 and 5 rounds. 
Conversely, after 5 rounds, the improvements become less pronounced, indicating diminishing returns. We observe an increase below 1.5\% for mAP50 and mAP50-95. Recall presents the higher increase between 5 and 8 rounds, namely 2.75\%. On the other hand, precision only shows an improvement of 0.37\%.
Concerning the training time, we note a slight difference between a lower number of rounds ($\leq5$) and more than 5 rounds. More precisely, between 3 and 5 rounds the increase time is 16.77 seconds while from 5 to 8 rounds the increase is around 18 seconds. However, the marginal performance gains from additional rounds may not justify the increased resource burden. These results suggest a trade-off between model performance and the computational costs.

Compared to baseline results, we notice a closer model performance, but this comes at the cost of computational resources. Namely, increasing the number of rounds to 8 with 200 epochs takes almost 5 times longer than the baseline.

\subsubsection{Aggregation Methods}
Focusing on the best results achieved previously, we compare the performance of FedAvg~\citep{mcmahan2017communication} and FedOpt~\citep{reddi2020adaptive}, two popular aggregation methods in FL. While FedAvg averages the model updates without considering learning dynamics, FedOpt introduces adaptive optimization techniques, which can adjust the learning rate during aggregation for potentially faster convergence.
Table \ref{tab:results} presents the comparative analysis of the performance of such aggregation methods over 8 rounds.

\begin{table*}[!htb]
\caption{Performance metrics results using FedAvg and FedOpt over different epochs and 8 rounds.}
\label{tab:results}
\centering
\scriptsize
\begin{tabular}{|c|c|c|c|c|c|c|c|}
\hline
Method   & Epochs & mAP50 & mAP50-95 & Recall  & Precision & Loss  & Training Time (sec) \\
\hline
   & 25     & 73.35\%  & 52.93\%   & 70.32\% & 84.36\%   & 0.00043 & 13,328 \\
   & 50     & 72.34\%  & 50.85\%   & 66.54\% & 76.01\%   & 0.00042 & 25,199 \\
FedOpt   & 100    & 75.11\%  & 52.42\%   & 66.05\% & 75.69\%   & 0.00041 & 48,591 \\
   & 150    & 73.62\%  & 51.49\%   & 67.86\% & 76.47\%   & 0.00040 & 57,734 \\
   & \underline{200}    & \underline{76.31\%}  & \underline{53.81\%}   & \underline{70.83\%} & \underline{79.14\%}   & \underline{0.00037} & \underline{68,205} \\
\hline
   & 25     & 71.27\%  & 52.23\%   & 67.34\% & 76.17\%   & 0.00041 & 13,307 \\
   & 50     & 72.48\%  & 50.93\%   & 68.49\% & 74.64\%   & 0.00040 & 25,176 \\
FedAvg   & 100    & 75.62\%  & 53.47\%   & 69.58\% & 80.26\%   & 0.00042 & 48,317 \\
   & 150    & 76.51\%  & 53.24\%   & 69.43\% & 73.75\%   & 0.00040 & 58,780 \\
   & \underline{200}    & \underline{76.69\%}  & \underline{53.57\%}   & \underline{71.27\%} & \underline{73.92\%}   & \underline{0.00039} & \underline{67,798} \\
\hline
\end{tabular}%

\end{table*}

In general, both methods demonstrate similar performance values. The maximum is reached at 200 epochs. However, FedOpt shows a consistent improvement of the loss metric over epochs, while FedAvg shows the same behavior for mAP50. Additionally, we observe that FedOpt presents a better mAP50-95, precision and loss.
In contrast, FedAvg presents better results for the remaining metrics.

Despite the high similarity between the two aggregation methods, we continue to experiment with FedOpt due to its superior optimization techniques, especially when dealing with potential variations in data quality between participants. FedOpt allows for more efficient convergence and provides stability in training. These characteristics make it more suitable for scenarios where the quality and availability of participant data may be uncertain.

\subsection{Anonymisation Layer}\label{sec:anonym}
\begin{figure}[!htb]
\begin{center}
    \includegraphics[width=1\linewidth]{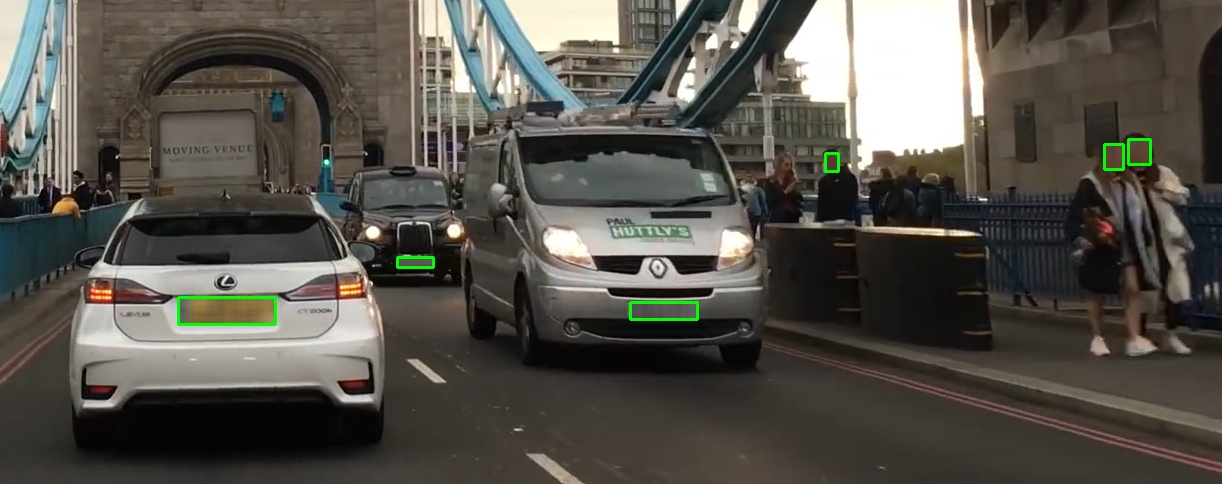}
\end{center}
   \caption{\small  A visual representation of the anonymization layer applied to an image. The green bounding boxes highlight the areas where anonymization was applied, specifically targeting license plates and faces.}
\label{img:anonym}
\end{figure}

After the best-federated setting selection, we demonstrate the effectiveness of the anonymization layer.
For this, we obfuscate the identified critical regions, namely faces and license plates using blurring through the Gaussian process.
Figure~\ref{img:anonym} illustrates an example of the anonymization layer in action. Key identifiable areas are highlighted with green bounding boxes, indicating the regions that were targeted for anonymization.%

While the primary goal of the anonymization layer is to protect privacy, it is also essential to maintain the data utility for downstream object detection. The anonymization process was carefully designed to minimize the loss of critical information that is necessary for model performance. For instance, while license plates are blurred, the overall structure and context of the vehicle are preserved, possibly allowing a model to still learn from the scene effectively.%

In summary, employing an object detection algorithm within a federated learning framework allows for a secure environment, where visual data remains stored locally in each organization's data repository. Although this results in some performance losses compared to the local training, i.e. centralized data, these losses are minimal with the advantage of avoiding higher privacy threats. As shown in Figure~\ref{img:anonym}, our method demonstrates a strong capability in detecting sensitive regions, with the anonymization layer effectively obfuscating this sensitive data, further enhancing privacy protection. 

\subsection{Discussion}\label{sec:discussion}
We focus our discussion on the challenges and implications of implementing and deploying an FL system. This includes the effects of scaling the number of participants, communication efficiency, and the scalability of FL models in real-world scenarios. Additionally, we address the ethical considerations associated with FL use, particularly regarding privacy, bias, and the risk of identification through unique visual features. This analysis aims to provide a deep understanding of the limitations and potential of FL, offering insights that can guide future research and practical development and deployment.

\paragraph{Expanding the number of participants.} 
Increasing the number of participants is expected to improve modelling performance, as more participants usually mean more data and subsequently better performance. However, for research purposes, we divide the dataset among all participants. This division results in each participant having access to a smaller subset of the data, which poses significant challenges for training robust and accurate models. Consequently, the performance of the federated model was lower compared to centralized training, where the model was trained on the entire dataset.

This suggests that while FL is a promising approach for enhancing privacy, its effectiveness is highly contingent on the availability of sufficient data per participant. In real-world applications, where data is distributed across many devices, strategies to mitigate the effects of data scarcity such as data augmentation, transfer learning, or federated data synthesis may be necessary to maintain model performance.

\paragraph{Communication costs and efficiency.}
Another critical factor in FL is the communication overhead between participants and the central server. At each round of training, participants are required to send model updates to the server, which then aggregates these updates and sends back the updated model parameters. As demonstrated in Table~\ref{tab:evolution_rounds}, there is a compromise between model performance and computational costs when increasing the rounds. Hence, this complexity increases with the introduction of a new participant to the FL environment, potentially leading to delays and increased energy consumption, especially in large-scale deployments.

This could be a bottleneck in scenarios where network bandwidth is limited or where participants operate on low-power devices, such as IoT sensors or mobile phones.
~\citet{wang2024horizontal} highlight this bottleneck by proposing the use of model compression techniques and asynchronous updates to address the presented challenges. The proposed methods might also be beneficial in our approach, which will be analyzed in future research.

\paragraph{Scalability and real-world deployment challenges.}
Previous concerns intensify the scalability problem.
For instance, increasing the number of participants introduces operational complexities such as participant dropout, variable participation rates, and inconsistent data availability~\citep{memia2023federated}.
Also, participants may join or leave the training process at any time, leading to fluctuations in model quality.~\citet{memia2023federated} highlight the importance of robust participant management strategies and potentially using fault-tolerant FL algorithms that can handle such variability without compromising model integrity.

Additionally, regulatory considerations, especially in sectors like healthcare and finance, add another layer of complexity to the adoption of this technology. FL must not only protect privacy through its decentralized approach but also comply with data protection regulations, such as GDPR. Thereby, ensuring that federated systems can both comply with legal frameworks and maintain technical integrity for a successful deployment.

\paragraph{Ethical considerations and bias mitigation.}
Despite FL being designed to ensure the protection of data privacy through a decentralized training process, the result models may still intensify biases present in the training data~\citep{mohri2019agnosticfederatedlearning}. This is particularly relevant in the context of visual data, where biases can result in the misidentification or inadequate anonymization of specific ethnic groups, potentially leading to unfair outcomes.
Future research should prioritize the development of bias mitigation techniques within FL frameworks to ensure that AI models do not inadvertently harm vulnerable populations.

Also, transparency in model development and deployment is essential for public trust. Open communication about how FL models are trained, including the types of data used and the measures taken to address bias, is essential to ensure ethical outcomes.

\paragraph{Identification through unique visual characteristics.}
Although the anonymization layer obfuscates sensitive regions, there may still be inherent risks of privacy leakage, such as reversing engineering~\citep{fredrikson2015model}. It is crucial to evaluate the possible threats in anonymized images. This challenge will be addressed in future work.
Most importantly, we highlight the potential risks associated with unique visual characteristics that may identify an individual even in anonymized images.
For instance, a person wearing a distinctive piece of clothing or having a unique tattoo may still be uniquely identifiable even after anonymization techniques have been applied. This concern is particularly relevant in applications, such as surveillance footage or medical images. 
As noted by~\citet{hukkelås2023does}, traditional anonymization methods may alleviate privacy concerns, but they might not fully prevent re-identification if unique visual features are not adequately obfuscated.

Thus, despite the obfuscation of critical regions as shown in Figure~\ref{img:anonym}, future research should include the development of advanced AI models to detect unique characteristics and automatically anonymize them.

\section{\uppercase{Conclusions}}
\label{sec:conclusion}

This paper presents the first framework that integrates object detection and Federated Learning (FL) with anonymization to enhance privacy in visual data management. By enabling decentralized model training, FL ensures privacy without sharing raw data, while YOLOv8 achieves efficient object detection of sensitive information. Our results demonstrate that despite some performance losses of the federated model compared to the baseline model (centralized), our solution is highly efficient in detecting critical regions such as faces and license plates. The detected regions are protected with an anonymization layer that effectively masks these identifiable features.
combining object detection, FL and anonymization techniques, provides a robust approach to secure visual data management, laying the foundation for future innovation in this field.

\bibliographystyle{apalike}
{\small
\bibliography{example}}

\end{document}